\pdfoutput=1
\documentclass[11pt]{article}

\usepackage[final]{acl}

\usepackage{times}
\usepackage{latexsym}

\usepackage[T1]{fontenc}
\usepackage[utf8]{inputenc}

\usepackage{microtype}

\usepackage{inconsolata}

\usepackage{graphicx}

\usepackage{amsmath}
\usepackage{amssymb}
\usepackage{mathtools}
\usepackage{amsthm}
\usepackage{enumitem}
\usepackage{multirow}
\usepackage{booktabs}
\usepackage{hyperref}

\def\huggingface{\raisebox{-1.5pt}{\includegraphics[height=1.05em]{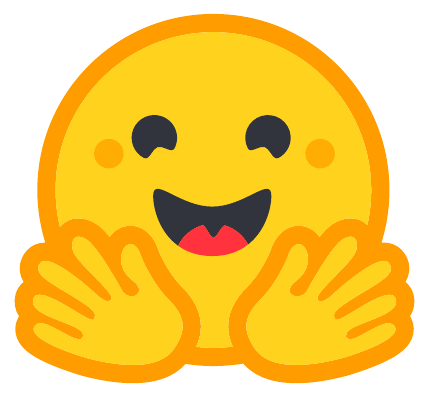}}}
\def\github{\raisebox{-1.5pt}{\includegraphics[height=1.0em]{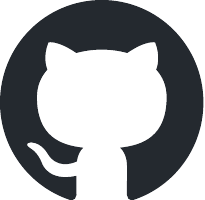}}}
\title{LLaVE: Large Language and Vision Embedding Models \\ with Hardness-Weighted Contrastive Learning}

\author{
 \textbf{Zhibin Lan\textsuperscript{1}}\thanks{~~Work was done when Zhibin Lan was interning at Pattern Recognition Center, WeChat AI, Tencent Inc, China.},\quad
 \textbf{Liqiang Niu\textsuperscript{2}},\quad
 \textbf{Fandong Meng\textsuperscript{2}},\quad
 \textbf{Jie Zhou\textsuperscript{2}},\quad
 \textbf{Jinsong Su\textsuperscript{1,3}}\thanks{~~Corresponding author.}
\\
 \textsuperscript{1}School of Informatics, Xiamen University, China,\\
 \textsuperscript{2}Pattern Recognition Center, WeChat AI, Tencent Inc, China, \\
  \textsuperscript{3}Shanghai Artificial Intelligence Laboratory, China \\
 \small{
   {lanzhibin@stu.xmu.edu.cn,\quad jssu@xmu.edu.cn}
 }\\
 \small{
 {\{poetniu, fandongmeng, withtomzhou\}@tencent.com}
 }\\
 \small{
 \github \enspace \href{https://github.com/DeepLearnXMU/LLaVE}{LLaVE} \quad \huggingface \enspace \href{https://huggingface.co/zhibinlan/LLaVE-0.5B}{LLaVE-0.5B} \quad 
 \huggingface \enspace \href{https://huggingface.co/zhibinlan/LLaVE-2B}{LLaVE-2B} \quad 
 \huggingface \enspace \href{https://huggingface.co/zhibinlan/LLaVE-7B}{LLaVE-7B}
 }
}

\begin{document}
\maketitle
\begin{abstract}
Universal multimodal embedding models play a critical role in tasks such as interleaved image-text retrieval, multimodal RAG, and multimodal clustering. However, our empirical results indicate that existing LMM-based embedding models trained with the standard InfoNCE loss exhibit a high degree of overlap in similarity distribution between positive and negative pairs, making it challenging to distinguish hard negative pairs effectively. To deal with this issue, we propose a simple yet effective framework that dynamically improves the embedding model's representation learning for negative pairs based on their discriminative difficulty. Within this framework, we train a series of models, named LLaVE, and evaluate them on the MMEB benchmark, which covers 4 meta-tasks and 36 datasets. Experimental results show that LLaVE establishes stronger baselines that achieve state-of-the-art (SOTA) performance while demonstrating strong scalability and efficiency. Specifically, LLaVE-2B surpasses the previous SOTA 7B models, while LLaVE-7B achieves a further performance improvement of 6.2 points. Although LLaVE is trained on image-text data, it can generalize to text-video retrieval tasks in a zero-shot manner and achieve strong performance, demonstrating its remarkable potential for transfer to other embedding tasks.
% \footnote{We release our source code and model upon the acceptance of
% our paper.}
\end{abstract}

\section{Introduction}
Multimodal embedding models aim to encode inputs from any modality into vector representations, which then facilitate various multimodal tasks, such as image-text retrieval \cite{DBLP:conf/cvpr/WuGGARGF21,DBLP:conf/icml/0033LHLQC0C24}, automatic evaluation \cite{clipscore}, and retrieval-augmented generation (RAG) \cite{DBLP:conf/emnlp/ZhaoCWJLQDGLLJ23}. Although advanced pretrained vision-language models such as CLIP \cite{CLIP}, ALIGN \cite{ALIGN}, and SigLIP \cite{SigLIP} can provide unified representations for text and images, they face difficulties when dealing with more complex tasks. Particularly, they adopt a dual-encoder architecture that encodes images and text separately, leading to poor performance in tasks such as interleaved image-text retrieval \cite{DBLP:conf/cvpr/WuGGARGF21}.

In recent years, the rapid development and exceptional performance of large multimodal models (LMMs) have prompted researchers to focus increasingly on LMM-based multimodal embedding models. Compared to traditional pretrained vision-language models, LMMs not only demonstrate superior multimodal semantic understanding capabilities but also naturally support interleaved text-image inputs \cite{gpt-4v,LLaVA,LLaVA1.5,lan-etal-2025-avg,DBLP:journals/corr/abs-2508-04453}. This advantage makes them more flexible and efficient in handling multimodal embedding tasks. As a representative work, \citeauthor{VLM2Vec} (\citeyear{VLM2Vec}) construct the Massive Multimodal Embedding Benchmark (MMEB), which encompasses 4 meta-tasks and 36 datasets, and train multimodal embedding models based on LMMs. Experiment results demonstrate that by providing suitable task instructions and employing contrastive learning for training, LMM can significantly outperform existing multimodal embedding models and generalize effectively across diverse tasks. 

\begin{figure}[t]
    \centering
    \includegraphics[width=\columnwidth]{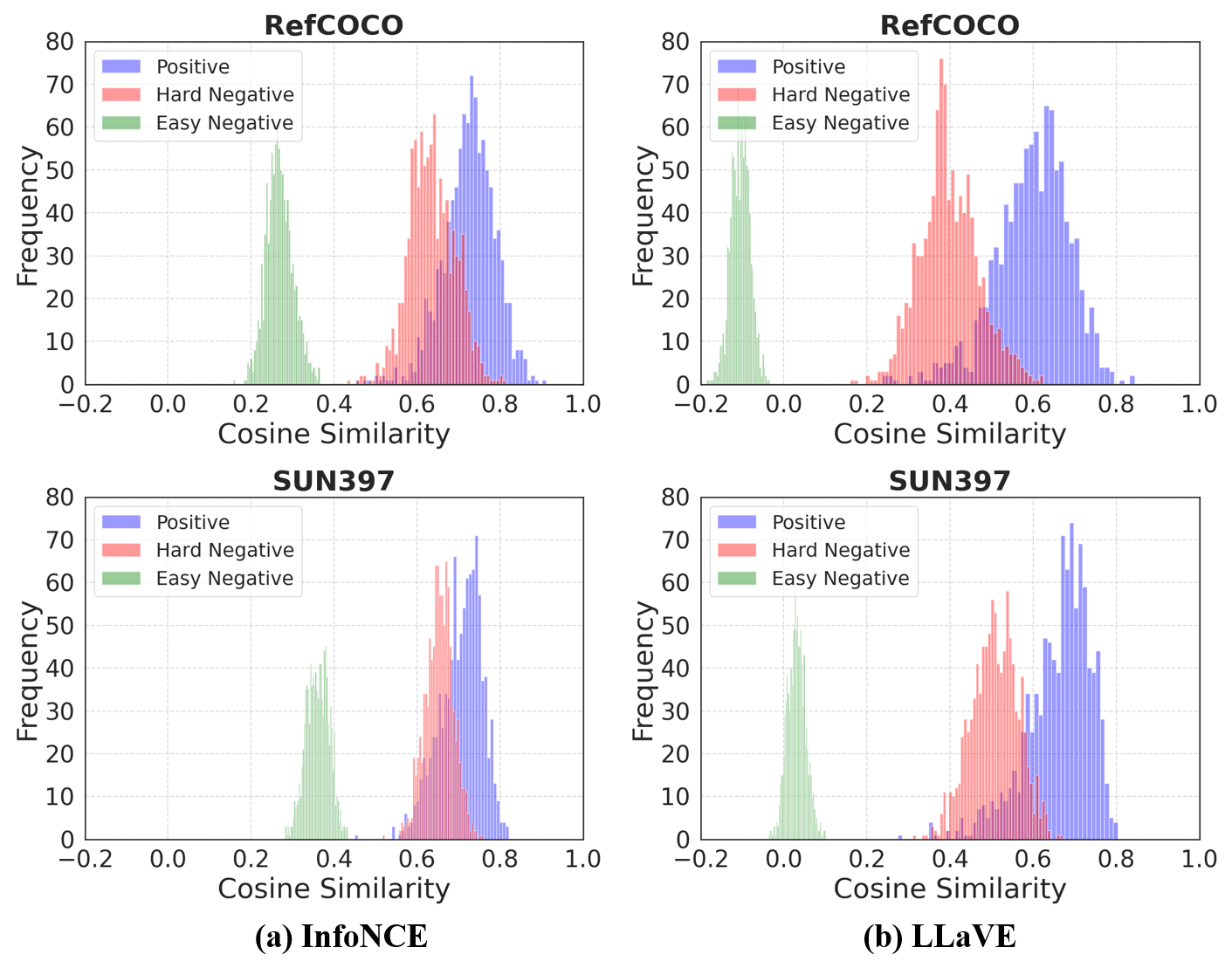}
    % \vspace{-0.8cm}
    \caption{Similarity distributions of learned embeddings on the SUN397 \cite{SUN397} and RefCOCO \cite{RefCOCO} dataset. We present the query-target cosine similarity histograms of positive, hard negative, and easy negative pairs for the model trained with the standard InfoNCE loss and LLaVE.}
    \label{figure:distribution}
% \vskip -0.1in
\end{figure}

However, as shown in Figure \ref{figure:distribution}(a), our preliminary study finds that when training an LMM as a multimodal embedding model using the standard InfoNCE loss \cite{InfoNCE}, the query-target similarity distribution of positive and negative pairs exhibits significant overlap. This indicates that the model struggles to learn discriminative multimodal representations for positive and hard negative pairs.

Building on the above observation and insights from preference learning \cite{DPO, DBLP:conf/aaai/00010LYHLW24}, we propose a simple yet effective framework to encourage the model to focus more on hard negative pairs, forcing it to learn more discriminative multimodal representations. Under our framework, we consider the embedding model as a policy model and introduce a reward model to assign an adaptive weight to each negative pair, where harder pairs are assigned with larger weights. This ensures that harder negative pairs play a more significant role in model training. In addition, by decoupling the reward model from the policy model, our framework can not only use different models for hardness estimation but also leverage manually annotated hardness to enhance the representation learning for specific samples. Inspired by SigLIP \cite{SigLIP}, we introduce a cross-device negative sample gathering strategy, which significantly alleviates the issue of limited negative samples in LMMs caused by excessive memory usage. As shown in Figure \ref{figure:distribution}, we observe that our framework increases the query-target similarity gap between positive and negative pairs, indicating its effectiveness in helping the model learn more discriminative multimodal representations.

To evaluate the effectiveness of our framework, we train a series of multimodal embedding models, referred to as LLaVE (Large Language and Vision Embedding Models), within the proposed framework. 
These models are based on advanced open-source LMMs of varying scales, including LLaVA-OV-0.5B \cite{LLaVA-OV}, Aquila-VL-2B \cite{DBLP:journals/corr/abs-2410-18558}, and LLaVA-OV-7B \cite{LLaVA-OV}. 
Experimental results on MMEB demonstrate that LLaVE-0.5B achieves comparable results to that of the previous VLM2Vec (phi-3.5-V-4B). When scaled up to LLaVE-2B, the model requires only about 17 hours of training on a single machine equipped with 8 A100 GPUs (40GB) to surpass the state-of-the-art (SOTA) model MMRet-7B, which is pretrained on 27 million image-text pairs. Furthermore, when expanded to LLaVE-7B, its performance is even more impressive, surpassing the previous SOTA model by 6.2 points. 
Meanwhile, when being scaled to different sizes, LLaVE still outperforms the models trained with InfoNCE loss based on the same LMM, demonstrating the effectiveness of our framework. 
These results fully validate that our framework is both easily scalable and resource-efficient. In addition, despite being trained exclusively on image-text data, LLaVE generalizes effectively to text-video retrieval tasks, showcasing its strong potential for transferring to other embedding tasks.

\section{Preliminary Study}
In this section, we first briefly formulate the multimodal embedding task and review the standard InfoNCE loss. Then, we analyze the similarity distributions of positive and negative pairs in LMM-based embedding models trained with the standard InfoNCE loss.

\begin{table*}[]
\centering
\setlength\tabcolsep{6.5pt}%调列距
\renewcommand\arraystretch{1.2}
\small
\fontsize{7pt}{7pt}\selectfont
% \resizebox{\textwidth}{!}{
\begin{tabular}{lcccccccc}
\toprule
\multicolumn{1}{l|}{\multirow{2}{*}{Type}} & \multicolumn{2}{c|}{Classification}            & \multicolumn{2}{c|}{VQA}                        & \multicolumn{2}{c|}{Retrieval}                 & \multicolumn{2}{c}{Visual Grounding} \\
\multicolumn{1}{l|}{}                      & SUN397      & \multicolumn{1}{c|}{ImageNet-R}  & DocVQA       & \multicolumn{1}{c|}{TextVQA}     & CIRR        & \multicolumn{1}{c|}{Wiki-SS-NQ}  & MSCOCO            & RefCOCO-Matching          \\ \midrule
\multicolumn{9}{c}{\textit{InfoNCE}}                                                                                                                                                                                                           \\ \midrule
\multicolumn{1}{l|}{Positive}              & 0.71        & \multicolumn{1}{c|}{0.68}        & 0.70          & \multicolumn{1}{c|}{0.69}        & 0.78        & \multicolumn{1}{c|}{0.57}        & 0.82              & 0.73             \\
\multicolumn{1}{l|}{Hard Negative}         & 0.65(-0.06) & \multicolumn{1}{c|}{0.59(-0.09)} & 0.62(-0.08)  & \multicolumn{1}{c|}{0.62(-0.07)} & 0.76(-0.02) & \multicolumn{1}{c|}{0.55(-0.02)} & 0.76(-0.06)       & 0.64(-0.09)      \\
\multicolumn{1}{l|}{Easy Negative}         & 0.36(-0.35) & \multicolumn{1}{c|}{0.38(-0.30)} & 0.30(-0.40)   & \multicolumn{1}{c|}{0.32(-0.37)} & 0.41(-0.37) & \multicolumn{1}{c|}{0.29(-0.28)} & 0.39(-0.43)       & 0.27(-0.46)      \\
\midrule
\multicolumn{1}{l|}{Precision@1 $\uparrow$}               & 66.2       & \multicolumn{1}{c|}{86.4}       & 81.2        & \multicolumn{1}{c|}{76.3}       & 39.7       & \multicolumn{1}{c|}{56.6}       & 76.1             & 83.8            \\ \midrule
\multicolumn{9}{c}{\textit{LLaVE}}                                                                                                                                                                                                              \\ \midrule
\multicolumn{1}{l|}{Positive}              & 0.66        & \multicolumn{1}{c|}{0.58}        & 0.61         & \multicolumn{1}{c|}{0.59}        & 0.63        & \multicolumn{1}{c|}{0.45}        & 0.62              & 0.93             \\
\multicolumn{1}{l|}{Hard Negative}         & 0.51(-0.15) & \multicolumn{1}{c|}{0.39(-0.19)} & 0.42(-0.19)  & \multicolumn{1}{c|}{0.45(-0.14)} & 0.56(-0.07) & \multicolumn{1}{c|}{0.38(-0.07)} & 0.52(-0.10)       & 0.64(-0.29)      \\
\multicolumn{1}{l|}{Easy Negative}         & 0.03(-0.63) & \multicolumn{1}{c|}{0.07(-0.51)} & -0.04(-0.65) & \multicolumn{1}{c|}{0.01(-0.58)} & 0.01(-0.62) & \multicolumn{1}{c|}{0.03(-0.42)} & -0.02(-0.64)      & 0.05(-0.88)      \\
\midrule
\multicolumn{1}{l|}{Precision@1 $\uparrow$}               & \textbf{75.5}       & \multicolumn{1}{c|}{\textbf{89.1}}       & \textbf{88.5}        & \multicolumn{1}{c|}{\textbf{78.8}}       & \textbf{50.0}         & \multicolumn{1}{c|}{\textbf{64.4}}       & \textbf{80.0}               & \textbf{85.5}            \\ 
\bottomrule
\end{tabular}
% }
% \vspace{-0.2cm}
\caption{The average cosine similarity between queries and targets is reported for positive, hard negative, and easy negative pairs across eight datasets. The numbers in parentheses represent the similarity difference between negative and positive pairs, where lower values indicate a smaller overlap between the two similarity distributions. It can be observed that our method effectively increases the similarity gap between negative and positive pairs, resulting in higher precision.}
\label{tab:preliminary}
\vskip -0.15in
\end{table*}

\subsection{Contrastive Learning for LMM-based Multimodal Embedding Models}
\label{sec:contrastive_learning}
Following VLM2Vec \cite{VLM2Vec}, we address the challenge of universal retrieval with LMMs. Specifically, given a query-target pair, it can be represented as ($q$, $t^+$), where both $q$ and $t^+$ could be an image, text, or interleaved image-text input. Note that $q$ will be equipped with corresponding task instructions for different tasks. The objective of this task is to ensure that the similarity between $q$ and $t^+$ is greater than the similarities between $q$ and other negative candidates $\{t^-\}$.

The aim of contrastive learning is to learn discriminative multimodal representations by pulling closer the representations of queries and targets in positive pairs while pushing apart the representations of queries and targets in negative pairs \cite{DBLP:conf/cvpr/HadsellCL06}. Given an LMM, we input the query and the target separately into the model and obtain their representations by extracting the vector representations of the last token in the final layer. Formally, for a mini-batch of training data $\{(q_1, t_1),..., (q_N, t_N)\}$, the standard InfoNCE loss is defined as
\begin{equation}
    \mathcal{L} = \frac{1}{N} \sum_{i=1}^{N} \underbrace{-log\frac{e^{{s_{i,i}/\tau}}}{e^{{s_{i,i}/\tau}}+\sum_{j\neq i}^{N}e^{{s_{i,j}/\tau}}}}_{\mathcal{L}_i},
\end{equation}
where $s_{i,j}=\mathrm{cosine}(\mathrm{LMM}(q_i), \mathrm{LMM}(t_i))$, $\mathrm{LMM}(\cdot)$ denotes the use of LMM for obtaining the representation, and $\tau$ represents the temperature hyper-parameter. In this work, $\tau$ is always set to 0.02 following the setting of VLM2Vec \cite{VLM2Vec}.

\subsection{Analysis}
We use Aquila-VL-2B \cite{DBLP:journals/corr/abs-2410-18558} as the base model, which builds upon the LLaVA-OneVision architecture, and perform contrastive learning on the MMEB dataset following the VLM2Vec setup. Then, we define the five pairs with the highest query-target similarities (excluding the positive pairs) as hard negative pairs, and the five pairs with the lowest similarities as easy ones. Subsequently, we use the cosine function to calculate the average query-target similarity for the two groups, respectively. As shown in the left part of Figure \ref{figure:distribution}, we visualize the similarity distributions of the trained model on SUN397 and RefCOCO. It can be observed that the distributions of positive and hard negative pairs exhibit significant overlap, while easy negative pairs also demonstrate relatively high similarities. 

To further explore, we randomly select one in-distribution dataset and one out-of-distribution dataset from the four meta-tasks in MMEB to evaluate the model's average query-to-target similarity on positive, hard negative, and easy negative pairs, respectively. As shown in Table \ref{tab:preliminary}, the similarity difference between positive and negative pairs in models trained with InfoNCE loss is relatively small (no more than 0.09), especially on CIRR and Wiki-SS-NQ, where the difference is only 0.02. Besides, the model exhibits the lowest precision on these two datasets. \textbf{Empirically, we observe that the smaller the similarity difference between positive and negative pairs, the lower the final precision tends to be}. This observation validates the necessity of enhancing learning on hard negative pairs during training, motivating us to explore a simple and effective approach to strengthen the model's learning of negative pairs with varying difficulty levels.

\section{Our Framework}
In this section, we first illustrate the inherent consistency between preference learning and contrastive learning. We then propose a simple yet
effective framework that incorporates hardness-weighted contrastive learning (See Section \ref{sec:hardness_weighted}) and cross-device negative sample gathering (See Section \ref{sec:cross_device}).

\subsection{Hardness-Weighted Contrastive Learning}
\label{sec:hardness_weighted}
Preference learning and contrastive learning share a fundamental goal: modeling relationships between pairs based on relative preference or similarity of the target within the pairs. Generally, preference learning involves a reward model and a policy model. The reward model scores the outputs of the policy model, while the policy model updates its parameters using the feedback from the reward model to produce higher-reward outputs. 

As a typical representative of preference learning, the Bradley-Terry (BT) model \cite{bradley1952rank} captures pairwise relationships through probabilistic comparisons. To directly optimize the embedding model (i.e. the policy model), we follow \citeauthor{DBLP:conf/aaai/00010LYHLW24} (\citeyear{DBLP:conf/aaai/00010LYHLW24}) to consider the embedding model as both the reward model and policy model. Formally, given a query $q_1$ and two targets $t_1$ and $t_2$, the BT model defines the training objective of preferring $t_1$ over $t_2$ as
\begin{equation}
    \mathcal{L}_1 = -\log\frac{e^{{r_{\pi}(q_1,t_1)}}}{e^{{r_{\pi}(q_1,t_1)}}+e^{{{r_{\pi}(q_1,t_2)}}}},
    \label{eq:bt}
\end{equation}
where, $r_{\pi}(\cdot)$ denotes the function of reward/policy model. Naturally, we can extend the Bradley-Terry \cite{bradley1952rank} model to a one-to-$N$ contrast setting \cite{DBLP:conf/aaai/00010LYHLW24}, which is essentially consistent with the InfoNCE loss. As a result, Equation \ref{eq:bt} is derived as
\begin{equation}
    \mathcal{L}_i = -\log\frac{e^{{r_{\pi}(q_i,t_i)}}}{e^{{r_{\pi}(q_i,t_i)}}+\sum_{j\neq i}^{N}e^{{{r_{\pi}(q_i,t_j)}}}},
\end{equation}

\begin{figure}[t]
    \centering
    \includegraphics[width=\columnwidth]{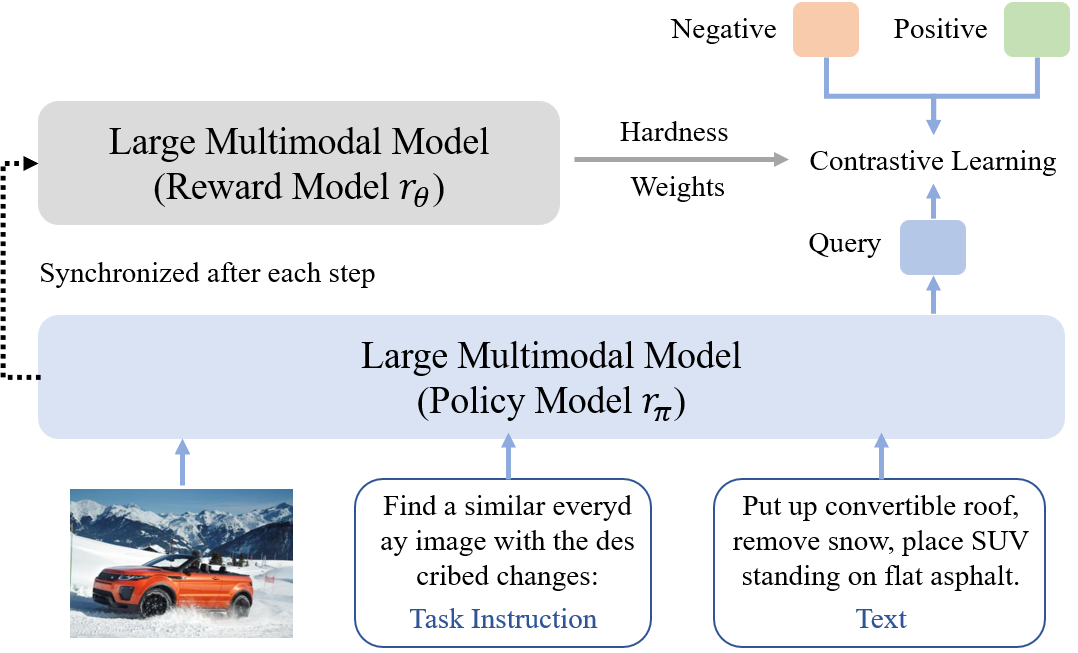}
    \vspace{-0.6cm}
    \caption{Overview of hardness-weighted contrastive learning. Please note that the policy and reward models are identical in our work. The dashed line indicates directly copying the parameters of the policy model to the reward model.}
    \label{figure:model}
\end{figure} 

where $r_{\pi}(q_i,t_j) = s_{i,j}/\tau$, $s_{i,j}$ and $\tau$ have been defined in Section \ref{sec:contrastive_learning}. Based on observations from the preliminary study, we propose hardness-weighted contrastive learning that assigns weight according to the learning difficulty of the negative pair. Higher weights indicate greater difficulty and incur heavier penalties, encouraging the model to learn more from challenging negative pairs. To this end, the training objective is revised as
\begin{equation}
    \mathcal{L}_i = -\log\frac{e^{{r_{\pi}(q_i,t_i)}}}{e^{{r_{\pi}(q_i,t_i)}}+\sum_{j\neq i}^{N}w_{ij} \cdot e^{{{r_{\pi}(q_i,t_j)}}}},
\end{equation}
where $w_{ij}$ represents the weight of learning difficulty. As shown in Figure \ref{figure:model}, to estimate the learning difficulty of pairs, we introduce a reward model $r_{\theta}$ and set $w_{ij}=e^{r_{\theta}(q_i,t_j)}$. Accordingly, the policy model adjusts its learning of different negative pairs based on the feedback from the reward model.
In this work, to achieve higher training efficiency and simpler implementation, we update the reward model $r_{\theta}$ to keep aligned with the policy model $r_{\pi}$ after each step. The reward model does not perform backpropagation, i.e., $r_{\theta}(q_i,t_j)$ = $\alpha \cdot \mathrm{sg}(s_{ij})$, where $\alpha$ is a hyperparameter and $\mathrm{sg}(\cdot)$ denotes the stop-gradient operation. 
When the reward model is set to be the same as the policy model, another advantage is that assigning a higher reward to a negative sample indicates that the policy model finds it more difficult to distinguish, thereby enhancing its learning on currently hard samples.
Note that $r_{\theta}$ can also adopt model structures other than the policy model. Finally, $\mathcal{L}_i$ is defined as
\begin{equation}
    \mathcal{L}_i = -\log\frac{e^{{r_{\pi}(q_i,t_i)}}}
    {e^{{r_{\pi}(q_i,t_i}}) 
    + \sum_{j\neq i}^{N} e^{({r_{\pi}(q_i,t_j) + r_{\theta}(q_i,t_j)})}}.
\end{equation}
We further analyze the gradients with respect to the $r_{\pi}(q_i,t_j)$ ($j \neq i$), which are formulated as
\begin{equation}
    \frac{\partial L_i}{\partial r_{\pi}(q_i,t_j)}=e^{r_{\theta}(q_i,t_j)} \cdot \frac{e^{{{r_{\pi}(q_i,t_j)}}}}{Z_i},
\label{eq:gradient}
\end{equation}
\vspace{-0.2cm}
\begin{equation}
    Z_i = e^{{r_{\pi}(q_i,t_i)}}+\sum_{j\neq i}^{N}e^{({{r_{\pi}(q_i,t_j) +r_{\theta}(q_i,t_j)}})}.
\end{equation}
From Equation \ref{eq:gradient}, we can observe that the gradients of the negative pairs are proportional to the product $r_{\theta}(q_i,t_j)$. which implies that the greater the learning difficulty of a negative pair, the more significant its role in the gradient update.

\subsection{Cross-Device Negative Sample Gathering}
\begin{figure}[t]
    \centering
    \includegraphics[width=\columnwidth]{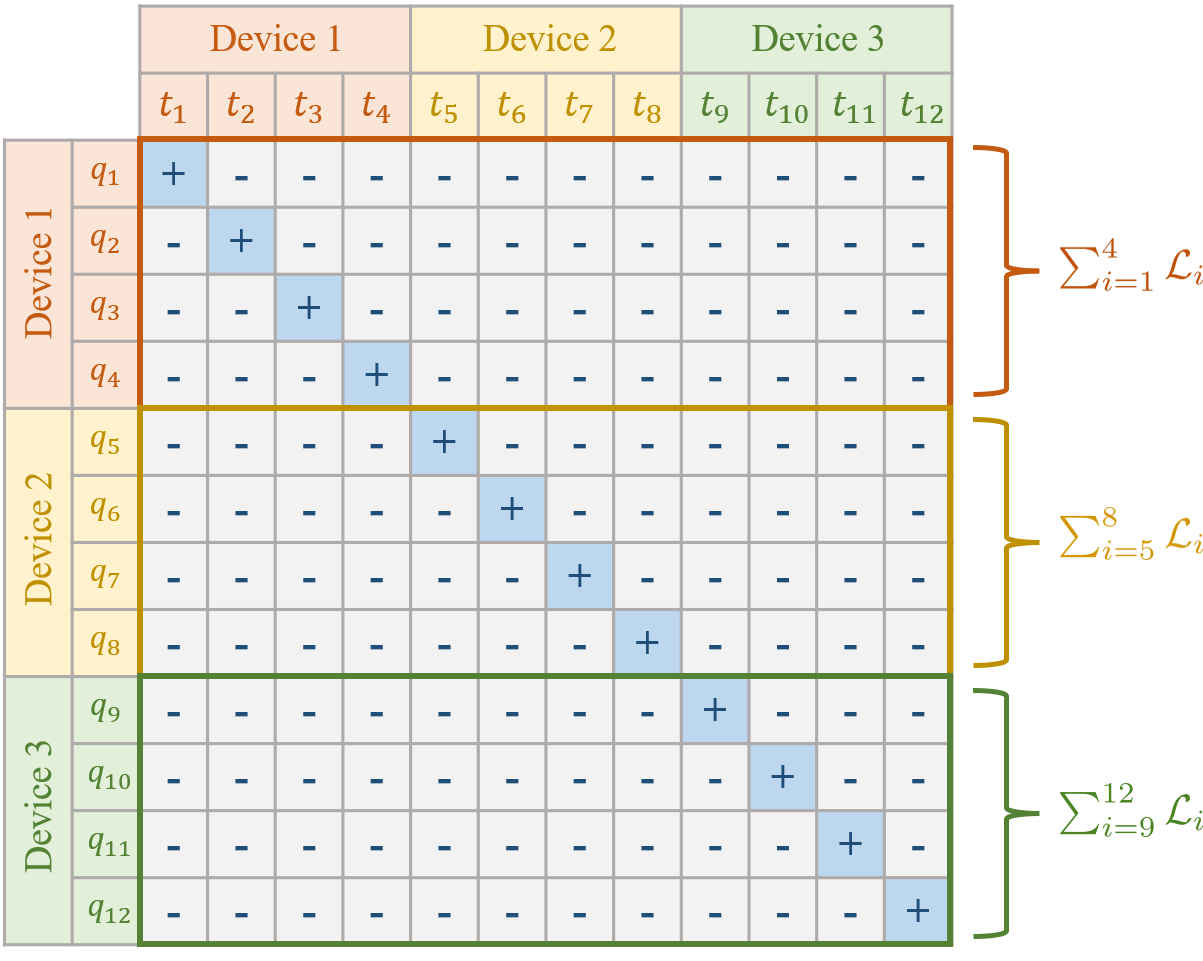}
    \vspace{-0.5cm}
    \caption{An example of cross-device negative sample gathering ($N$=$4$ and $K$=$3$). The plus signs represent positive pairs, and the minus signs represent negative pairs. Each device calculates the similarity between its own queries and the targets on all other devices, which is then used for loss computation.}
    \label{figure:cross-device}
% \vskip -0.2in
\end{figure} 
\label{sec:cross_device}
The number of negative pairs in contrastive learning has an important effect on model training. However, LMM-based embedding models face the challenge of high memory consumption, making it difficult to use a large batch size directly. To alleviate this issue, inspired by OpenCLIP \cite{DBLP:conf/cvpr/ChertiBWWIGSSJ23} and SigLIP \cite{SigLIP}, we adopt a cross-device negative sample gathering strategy, which increases the number of negative pairs by a factor of the device number $K$. As illustrated in Figure \ref{figure:cross-device}, we expand the number of negative pairs on each device by gathering samples from other devices. Consequently, $\mathcal{L}_i$ is reformulated as follows:
\begin{equation}
    \mathcal{L}_i = -\log\frac{e^{{r_{\pi}(q_i,t_i)}}}
    {e^{{r_{\pi}(q_i,t_i)}}
    + \sum_{j\neq i}^{N \cdot K} e^{({r_{\pi}(q_i,t_j) + r_{\theta}(q_i,t_j)})}}.
\end{equation}
With this strategy, we can effectively increase the number of negative pairs without significantly increasing memory consumption.

\section{Experiments}

\subsection{Setup}

\paragraph{Datasets and Metrics.} 
In this study, we follow VLM2Vec \cite{VLM2Vec} to train our model on 20 in-distribution datasets from MMEB. These datasets encompass four meta-tasks: classification, VQA, multimodal retrieval, and visual grounding, with a total of 662K training pairs. The model is then evaluated on both 20 in-distribution and 16 out-of-distribution test sets from MMEB. We report Precision@1 on each dataset, which measures the proportion of top-ranked candidates that are positive samples.

\begin{table*}[t]
\centering
\setlength\tabcolsep{6pt}%调列距
\renewcommand\arraystretch{1.2}
\small
\fontsize{8pt}{8pt}\selectfont
% \resizebox{\textwidth}{!}{
\begin{tabular}{lccccccc}
\toprule
\multicolumn{1}{c}{\multirow{2}{*}{\textbf{Model}}} & \multicolumn{4}{c}{\textbf{Per Meta-Task Score}}               & \multicolumn{3}{c}{\textbf{Average Score}}    \\ \cline{2-8} 
\multicolumn{1}{c}{}                                & Classification & VQA           & Retrieval     & Grounding     & IND           & OOD           & Overall       \\ \midrule
\# Datasets                                         & 10             & 10            & 12            & 4             & 20            & 16            & 36            \\ \midrule
\rowcolor[HTML]{EDEDED}
\multicolumn{8}{c}{\textit{Baselines}}                                                                                                                                         \\ \midrule
CLIP \cite{CLIP}                                                & 42.8           & 9.1           & 53.0          & 51.8          & 37.1          & 38.7          & 37.8          \\
BLIP2 \cite{BLIP2}                                              & 27.0           & 4.2           & 33.9          & 47.0          & 25.3          & 25.1          & 25.2          \\
SigLIP \cite{SigLIP}                                             & 40.3           & 8.4           & 31.6          & 59.5          & 32.3          & 38.0          & 34.8          \\
OpenCLIP \cite{DBLP:conf/cvpr/ChertiBWWIGSSJ23}                                            & 47.8           & 10.9          & 52.3          & 53.3          & 39.3          & 40.2          & 39.7          \\
UniIR (BLIP$_{FF}$) \cite{Uniir}                               & 42.1           & 15.0          & 60.1          & 62.2          & 44.7          & 40.4          & 42.8          \\
UniIR (CLIP$_{SF}$) \cite{Uniir}                                & 44.3           & 16.2          & 61.8          & 65.3          & 47.1          & 41.7          & 44.7          \\
Magiclens \cite{Magiclens}                                          & 38.8           & 8.3           & 35.4          & 26.0          & 31.0          & 23.7          & 27.8          \\ \midrule
\rowcolor[HTML]{EDEDED}
\multicolumn{8}{c}{\textit{LMM-based Baselines}}                                                                                                                              \\ \midrule
E5-V \cite{E5-V}                                               & 21.8           & 4.9           & 11.5          & 19.0          & 14.9          & 11.5          & 13.3          \\
VLM2Vec (Phi-3.5-V-4B) \cite{VLM2Vec}                                     & 54.8           & 54.9          & 62.3          & 79.5          & 66.5          & 52.0          & 60.1          \\
VLM2Vec (LLaVA-NeXT-7B-LR) \cite{VLM2Vec}                                & 54.7           & 50.3          & 56.2          & 64.0          & 61.0          & 47.5          & 55.0          \\
VLM2Vec (LLaVA-NeXT-7B-HR) \cite{VLM2Vec}                                & 61.2           & 49.9          & 67.4          & 86.1          & 67.5          & 57.1          & 62.9          \\
MMRet (LLaVA-NeXT-7B) \cite{MMRet}                                     & 56.0           & 57.4          & \underline{69.9}          & 83.6          & 68.0          & 59.1          & 64.1          \\ \midrule
\rowcolor[HTML]{EDEDED}
\multicolumn{8}{c}{\textit{Our trained LMM-based Baselines}}  \\
\midrule
VLM2Vec (LLaVA-OV-0.5B)                                & 54.6           & 44.7          & 56.8          & 76.5          & 59.8          & 49.1          & 55.0          \\
VLM2Vec (Aquila-VL-2B)                                & 61.1           & 57.3          & 62.1          & 85.5          & 67.2          & 58.1          & 63.1          \\
VLM2Vec (LLaVA-OV-7B)                                & \underline{63.5}           & \underline{61.1}          & 64.5          & \underline{87.3}          & \underline{69.7}          & \underline{61.0}          & \underline{65.8}          \\
\midrule
\rowcolor[HTML]{EDEDED}
\multicolumn{8}{c}{\textit{Ours}}                                                                                                                                             \\ \midrule
LLaVE-0.5B                                          & 57.4           & 50.3          & 59.8          & 82.9          & 64.7          & 52.0          & 59.1          \\
LLaVE-2B                                            & 62.1           & 60.2          & 65.2          & 84.9          & 69.4          & 59.8          & 65.2          \\
LLaVE-7B                                            & \textbf{65.7}  & \textbf{65.4} & \textbf{70.9} & \textbf{91.9} & \textbf{75.0} & \textbf{64.4} & \textbf{70.3} \\
\rowcolor[HTML]{FFFACD}
$\triangle$ - Best baseline                                            & +2.2  & +4.3 & +1.0 & +4.6 & +5.3 & +3.4 & +4.5 \\

\bottomrule
\end{tabular}
% }
\vspace{-0.2cm}
\caption{Results on the MMEB benchmark. IND represents the in-distribution dataset, and OOD represents the out-of-distribution dataset. In UniIR, the FF and SF subscripts under CLIP or BLIP represent feature-level fusion and score-level fusion, respectively. LLaVA-NeXT-7B-LR indicates the use of low-resolution (336×336) image inputs, while LLaVA-NeXT-7B-HR refers to the use of high-resolution (1344×1344) image inputs. The reported scores are the average Precision@1 over the corresponding datasets. The best results are marked in bold, and the second-best results are underlined. Part of the baseline results are sourced from \cite{VLM2Vec} and \cite{MMRet}. Detailed results and qualitative evaluations can be found in Appendix \ref{sec:detailed_result} and Section \ref{sec:qualitative}.}
\label{tab:main_results}
% \vskip -0.15in
\end{table*}

\paragraph{Implementation Details.} 
Our trained model, LLaVE, includes three scales: 0.5B, 2B, and 7B, based on LLaVA-OV-0.5B \cite{LLaVA-OV}, Aquila-VL-2B \cite{DBLP:journals/corr/abs-2410-18558}, and LLaVA-OV-7B \cite{LLaVA-OV}, respectively. To facilitate community use, the training code for LLaVE is built on the widely-used Transformers \cite{Transformers} and DeepSpeed \cite{Deepspeed} packages. We use a batch size of 256 by gradient accumulation, set the weighting hyperparameter $\alpha$ to 9 \footnote{We empirically analyze the impact of $\alpha$ on model performance in Appendix \ref{sec:hyper_analysis}.}, and impose a total length limit of 4096. Furthermore, we employ the Higher Anyres technique \cite{LLaVA-OV} to support high-resolution images, setting the maximum image resolution to 672 $\times$ 672. The learning rate is set to 1e-5 for LLaVE-0.5B and LLaVE-2B, and 5e-6 for LLaVE-7B.
For efficient training, we freeze the vision encoder and train the model for one epoch using the DeepSpeed ZeRO-3 strategy. Regarding training costs, LLaVE-0.5B and LLaVE-2B are trained on 8 NVIDIA A100 GPUs (40GB) for 12 and 17 hours, respectively, while LLaVE-7B is trained on 16 Ascend 910B GPUs (64GB) for 33 hours. More details can be found in Appendix \ref{sec:training_details}.

\paragraph{Baselines.} 
Following VLM2Vec, we compare our model with CLIP \cite{CLIP}, OpenCLIP \cite{DBLP:conf/cvpr/ChertiBWWIGSSJ23}, BLIP2 \cite{BLIP2}, SigLIP \cite{SigLIP}, UniIR \cite{Uniir}, E5-V \cite{E5-V}, and Magiclens \cite{Magiclens}. Additionally, we include two powerful models: VLM2Vec \cite{VLM2Vec} and MMRet-MLLM \cite{MMRet}. Among them, MMRet-MLLM enhances downstream task performance through pretraining on a self-built retrieval dataset consisting of 26 million pairs. To ensure a fairer comparison, we also compare the VLM2Vec trained using the same base LMM, including VLM2Vec (LLaVA-OV-0.5B), VLM2Vec (Aquila-VL-2B), and VLM2Vec (LLaVA-OV-7B).

\begin{table*}[]
\centering
\setlength\tabcolsep{11pt}%调列距
\renewcommand\arraystretch{1.2}
\small
\fontsize{8pt}{8pt}\selectfont
% \resizebox{\textwidth}{!}{
\begin{tabular}{lllll}
\toprule
ID & Model                                                                  & IND  & OOD  & Overall \\ \midrule
0  & Previous SOTA (MMRet)                                                  & 68.0 & 59.1 & 64.1    \\ \midrule
1  & Aquila-VL-2B + InfoNCE                                                 & 60.6 & 56.4 & 58.7    \\ \midrule
2  & 1 + Freeze image encoder                                                & 60.5 (-0.1) & 58.3 (+2.1) & 59.5 (+0.8)    \\
3  & 2 + Freeze projector                                                    & 60.3 (-0.2) & 57.0 (-1.3) & 58.8 (-0.7)   \\ \midrule
4  & 2 + Less training data (Each training dataset samples up to 20K data)  & 60.4 (-0.1) & 56.4 (-1.9) & 58.6 (-0.9)   \\
5  & 2 + More training data (Each training dataset samples up to 100K data) & 61.2 (+0.7) & 57.4 (-0.9) & 59.5 (+0.0)   \\ \midrule
6  & 2 + Cross-device negative sample gathering                                & 68.6 (+8.1) & 58.4 (+0.1) & 64.0 (+4.5)   \\
7  & 6 + Focal-InfoNCE loss \cite{DBLP:conf/emnlp/HouL23}                 & 67.9 (-0.7) & 59.5 (+1.1) & 64.2 (+0.2)   \\
8  & 6 + Hardness-weighted contrastive learning (LLaVE-2B)                  & 69.4 (+0.8) & 59.8 (+1.4) & 65.1 (+1.1)   \\ \bottomrule
\end{tabular}
% }
\vspace{-0.2cm}
\caption{Ablation results on MMEB. The model with ID 1 is configured to use the standard InfoNCE for full fine-tuning, with each training dataset sampling up to 50K examples, so as to ensure balance across different datasets. The numbers in parentheses indicate the impact of the changes on performance compared to the model corresponding to the previously selected ID.}
\label{tab:ablation}
\end{table*}

\subsection{Main Results}

Table \ref{tab:main_results} presents the performance comparison of our proposed LLaVE series (LLaVE-0.5B, LLaVE-2B, LLaVE-7B) against existing baseline models. Among the baseline models, our trained VLM2Vec (LLaVA-OV-7B) achieves the highest overall average score of 65.8, surpassing the current state-of-the-art model, MMRet, which achieves the second-best score of 64.1. This indicates that a more powerful foundational LMM can lead to better performance in the embedding models. Notably, MMRet excels in retrieval tasks with a score of 69.9, which is attributed to its pretraining on its self-constructed 26M image-text retrieval dataset. VLM2Vec (LLaVA-NeXT-7B-HR) exhibits superior performance in grounding tasks, likely due to its higher input image resolution, which achieves a 22.1-point improvement over VLM2Vec (LLaVA-NeXT-7B-LR).

Although previous models have achieved strong performance, our LLaVE series demonstrates consistent improvements over the best baseline across all metrics. LLaVE-7B achieves a state-of-the-art overall score of 70.3, outperforming the previous SOTA model, MMRet, by 6.2 points and surpassing VLM2Vec (LLaVA-OV-7B) by 4.5 points. In grounding, LLaVE-7B attains an impressive score of 91.9, a +4.6 point improvement over VLM2Vec. LLaVE-7B also leads in VQA (65.4) and classification (65.7), with improvements of +4.3 and +2.2 points, respectively. In addition, we observe that the performance of LLaVE models scales consistently with model size, indicating that our framework has excellent scalability. It is worth mentioning that the performance of LLaVE-0.5B is already comparable to VLM2Vec (Phi-3.5-V-4B), while LLaVE-2B achieves an overall score of 65.2, surpassing the previously pretrained MMRet-7B that utilizes an additional 27M image-text pairs.

\subsection{Ablation Study}
Table \ref{tab:ablation} provides an ablation study analyzing the impact of various design choices on the performance of LLaVE-2B across IND, OOD, and overall metrics. The baseline model (ID 1) uses the standard InfoNCE with up to 50K samples per training dataset for balanced fine-tuning.

\paragraph{Freezing the image encoder helps generalize to out-of-distribution datasets.} Freezing the image encoder (1 $\rightarrow$ 2) can significantly improve performance on unseen datasets at the cost of a slight decrease in in-distribution performance. This is likely because the original vision encoder possesses stronger generalization capabilities, and instruction fine-tuning on smaller datasets may affect this ability. However, freezing the projector (1 $\rightarrow$ 3) harms the model's performance because transforming the LMM into an embedding model requires re-adaptation.

\paragraph{When the data is sufficient, a balanced distribution of various data is more important than simply having more data.} Table \ref{tab:ablation} (2 $\rightarrow$ 4) demonstrates that with limited data, the model's generalization ability is constrained. When the data sampling limit is increased to 100K (2 $\rightarrow$ 5), certain training datasets expand further, while others remain unchanged due to limited availability. The results show that although this approach improves in-distribution performance, it negatively impacts generalization, highlighting the importance of maintaining a balanced distribution across different meta-tasks.

\paragraph{The number of negatives is crucial for training LMM-based embedding models.} By analyzing the performance of the model (ID 6), we observe that the introduction of the cross-device negative sample gathering strategy leads to substantial gains in IND (+8.1) with negligible impact of OOD (+0.1), resulting in a notable overall improvement of +4.5, which highlights the importance of diverse negative samples.

\paragraph{Hardness-weighted contrastive learning can further enhance the performance of powerful models on both in-distribution and out-of-distribution datasets.} Although the model (ID 6) already achieves near-SOTA performance, hardness-weighted contrastive learning further enhances its effectiveness, particularly with a 1.4-point improvement on out-of-distribution datasets, demonstrating its complementarity. Besides, we also compare the Focal-InfoNCE loss (ID 7), which also weights positive and negative pairs. Although it slightly improves performance on the out-of-distribution dataset, it reduces performance on the in-distribution dataset.
As shown in Figure \ref{figure:distribution} and Table \ref{tab:preliminary}, our framework significantly increases the similarity gap between positive and negative pairs, thereby improving the model's discriminative capability.

\begin{table}[]
\resizebox{0.5\textwidth}{!}{
\centering
\begin{tabular}{lcccccc}
\toprule
\multirow{2}{*}{\textbf{Model}} & \multicolumn{3}{c}{\textbf{MSR-VTT}}          & \multicolumn{3}{c}{\textbf{MSVD}}             \\
                                & R@1           & R@5           & R@10          & R@1           & R@5           & R@10          \\ \midrule
\multicolumn{7}{c}{\textit{Zero-shot (finetuned with text-video data)}}                                                                  \\ \midrule
InternVideo                     & \textcolor{gray}{40.0}          & \textcolor{gray}{65.3}          & \textcolor{gray}{74.1}          & \textcolor{gray}{43.4}          & \textcolor{gray}{69.9}          & \textcolor{gray}{79.1}          \\
ViCLIP                          & \textcolor{gray}{42.4}          & \textcolor{gray}{-}             & \textcolor{gray}{-}             & \textcolor{gray}{49.1}          & \textcolor{gray}{-}             & \textcolor{gray}{-}             \\
UMT-L                           & \textcolor{gray}{42.6}          & \textcolor{gray}{64.4}          & \textcolor{gray}{73.1}          & \textcolor{gray}{49.9}          & \textcolor{gray}{77.7}          & \textcolor{gray}{85.3}          \\
InternVideo2-6B                    & \textcolor{gray}{55.9}          & \textcolor{gray}{78.3}          & \textcolor{gray}{85.1}          & \textcolor{gray}{59.3}          & \textcolor{gray}{84.4}          & \textcolor{gray}{89.6}          \\ \midrule
\multicolumn{7}{c}{\textit{Zero-shot (finetuned only with text-image data)}}                                                             \\ \midrule
VLM2Vec                           & 43.5          & 69.3          & 78.9          & 49.5          & 77.5          & 85.7          \\
LamRA                           & 44.7          & 68.6          & 78.6          & 52.4          & 79.8          & 87.0          \\
LLaVE-7B        & \textbf{46.8} & \textbf{71.1} & \textbf{80.0} & \textbf{52.9} & \textbf{80.1} & \textbf{87.0} \\ \bottomrule
\end{tabular}
}
\caption{Results of zero-shot text-to-video retrieval. The gray font indicates that the model is trained using contrastive learning on tens of millions of text-video data.}
% \vskip -0.2in
\end{table}

\begin{figure*}[t]
% \vskip 0.2in
    \centering
    \includegraphics[width=2.0\columnwidth]{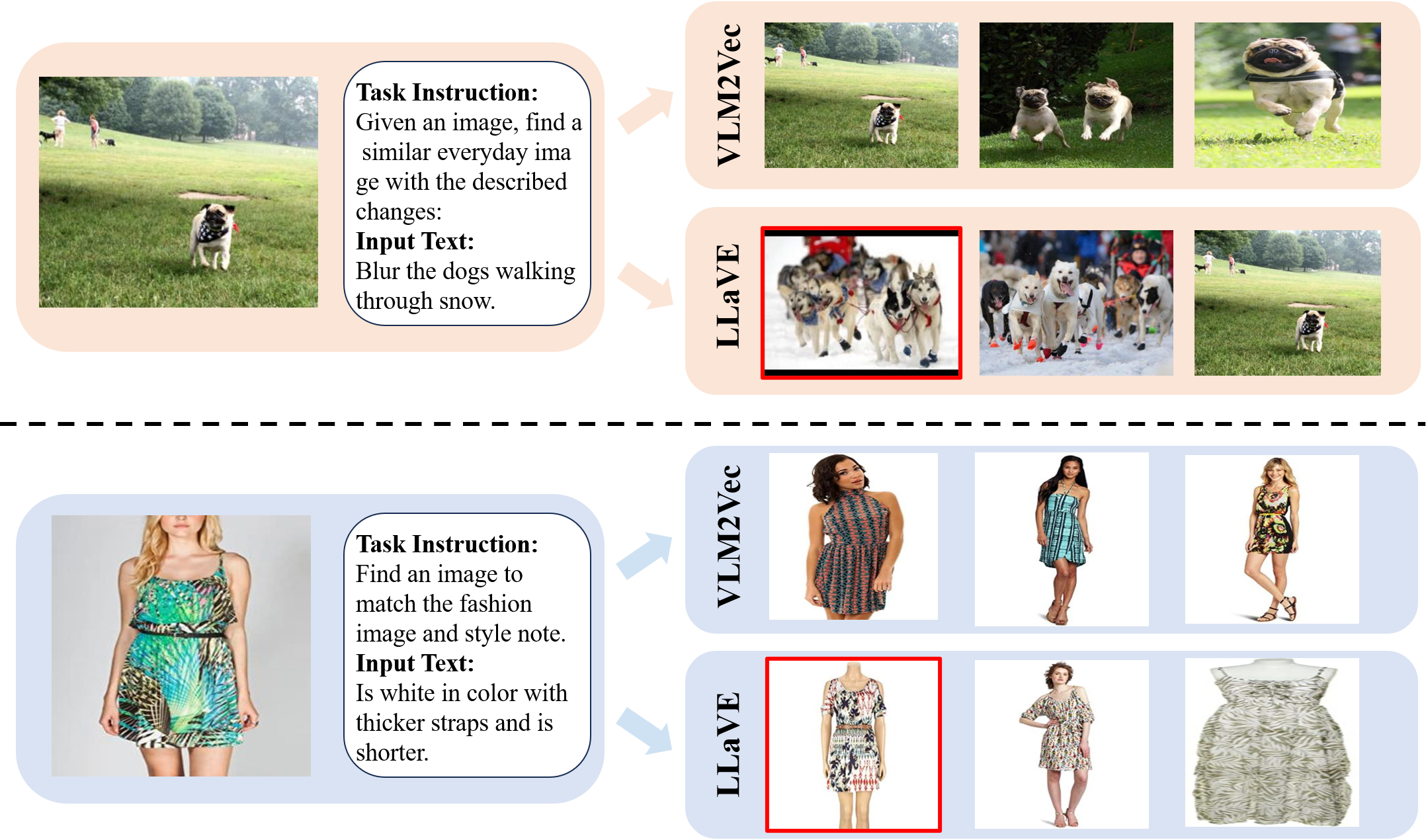}
    \caption{Qualitative evaluation comparing  LLaVE and VLM2Vec. Retrievals consistent with the ground truth are highlighted with red borders. From left to right, the images represent the top-1 to top-3 retrieval results.}
    \label{figure:case}
% \vskip -0.2in
\end{figure*} 

\subsection{Zero-shot Video Retrieval}
We also evaluate LLaVE on the widely used text-video retrieval datasets: MSR-VTT \cite{Msr-vtt} and MSVD \cite{MSVD}, to explore its generalization capability. There are two types of comparative models considered. The first type includes models trained on tens of millions of video-text data, such as InternVideo \cite{Internvideo}, ViCLIP \cite{Internvid}, UMT-L \cite{UMT-L}, and InternVideo2-6B \cite{InternVideo2-6B}. The second type is completely zero-shot, trained only on text-image data using contrastive learning and directly evaluated on text-video data. Particularly, we focus on the two strongest models, VLM2Vec (LLaVA-OV-7B) and LamRA \cite{LamRA}, which is based on Qwen2-VL-7B and consists of two 7B models: LamRA-Ret and LamRA-Rank. LamRA-Ret retrieves the top-K candidates, while LamRA-Rank further re-ranks these retrieved candidates. 

To enable video embedding, we set the maximum number of sampled frames to 32, expand the total input length of the model to 8192, and reduce the visual features of the video by 4 times through bilinear interpolation. It is observed that, compared to LamRA, LLaVE-7B requires only a single model and shows consistent improvements across all metrics. Notably, on MSR-VTT, the R@1, R@5, and R@10 scores increase by 2.1, 2.5, and 1.4, respectively. Moreover, although LLaVE-7B does not utilize text-video data for contrastive training, its performance still surpasses most video retrieval models except for InternVideo2-6B, which are trained on tens of millions of video-text pairs. These results demonstrate that LLaVE-7B has strong potential for transferring to other embedding tasks.

\subsection{Qualitative Evaluation}
\label{sec:qualitative}

In Figure \ref{figure:case}, we present the qualitative evaluation results of LLaVE and the strongest baseline model, VLM2Vec (LLaVA-OV-7B). As shown in the upper part of the figure, LLaVE successfully identifies and retrieves the modified target images that meet the specified requirement (``\textit{dogs walking through snow}") in the Top-2 retrieval, while VLM2Vec only retrieves images similar to the original image. Similarly, in the lower part of the figure, LLaVE fulfills the requirement of ``\textit{white in color and is shorter}" in the Top-3 retrieval. These examples demonstrate that our framework effectively facilitates the model to capture complex intents in challenging samples and enhances the discriminability of hard samples.

\section{Related Work}

\paragraph{Multimodal Embeddings.}
As a significant research direction, multimodal embeddings aim to integrate the information from multiple modalities (e.g., vision and language) into a shared representation space, which enables seamless understanding across modalities.
Early research primarily focuses on leveraging dual-stream architectures to separately encode texts and images. For instance, CLIP \cite{CLIP}, ALIGN \cite{ALIGN}, BLIP \cite{BLIP}, and SigLIP \cite{SigLIP} all adopt dual-encoder frameworks, learning universal representations from large-scale weakly supervised image-text pairs through contrastive learning. To learn the more universal multimodal representations, UniIR \cite{Uniir} proposes two fusion mechanisms to combine the visual and textual representations generated by the dual-encoder model. Although these models have achieved impressive results, they still face challenges in handling tasks such as interleaved image-text retrieval \cite{DBLP:conf/cvpr/WuGGARGF21} and instruction-following multimodal retrieval.

\paragraph{LMM-based Multimodal Embeddings.}
To address the aforementioned issue, E5-V \cite{E5-V} and VLM2Vec \cite{VLM2Vec} transform LMM into multimodal embedding models through contrastive learning, fully leveraging LMM's powerful multimodal understanding capability and its inherent advantage in handling interleaved text-image input. Recently, a few concurrent studies \cite{DBLP:conf/cvpr/ZhangZXLDLXZLZ25,DBLP:journals/corr/abs-2504-17432,DBLP:journals/corr/abs-2505-11293} further explored the application of LMMs in multimodal embeddings. For example, LamRA \cite{LamRA} adopts the retrieval model to select the top-K candidates, which are then scored by the reranking models. Finally, the scores from the retrieval and reranking models are combined using a weighted sum to produce the final score for retrieval. MMRet \cite{MMRet} creates a large-scale multimodal instruction retrieval dataset called MegaPairs. By pretraining on this dataset, MMRet achieves SOTA results on MMEB. 

\paragraph{Contrastive Learning.} Contrastive learning enables models to learn effective representations by distinguishing between positive and negative samples, and it has been widely applied across various domains \cite{DBLP:conf/iclr/KipfPW20,DBLP:conf/sigir/WuWF0CLX21,DBLP:journals/jair/LanLSXLWL23,DBLP:journals/ai/YinZSZMZHL23,DBLP:conf/emnlp/ZhangYFLSLMZWS24}.
Negative samples play a crucial role in contrastive learning, \citeauthor{DBLP:conf/icml/ChenK0H20} (\citeyear{DBLP:conf/icml/ChenK0H20}) show that incorporating more negative samples can enhance the performance of contrastive learning. \citeauthor{DBLP:conf/icml/AwasthiDK22} (\citeyear{DBLP:conf/icml/AwasthiDK22}) further explore the impact of the number of negative samples from both theoretical and empirical perspectives. Moreover, \citeauthor{DBLP:journals/corr/abs-2010-06682} (\citeyear{DBLP:journals/corr/abs-2010-06682}) demonstrate that hard negative samples are both necessary and sufficient to learn more discriminative representation, and \citeauthor{DBLP:conf/iclr/RobinsonCSJ21} (\citeyear{DBLP:conf/iclr/RobinsonCSJ21}) propose a hard negative sampling strategy where the user can control the hardness. The study most similar to ours is Focal-InfoNCE \cite{DBLP:conf/emnlp/HouL23}, which weighs both positive and negative pairs based on their query-target similarities. Specifically, it uses a fixed threshold to determine the hardness of negative pairs, increasing the weight if the similarity exceeds the threshold and decreasing it otherwise. Unlike this work, our hardness-weighted contrastive learning introduces a reward model to dynamically estimate the hardness of negative pairs and applies weighting only to the negative pairs based on the estimated hardness. Notably, the reward model can be decoupled from the policy model.

\section{Conclusion}
In this paper, we conduct a preliminary study to find that LMM-based embedding models trained with the standard InfoNCE loss face significant challenges in handling hard negative pairs. To address this issue, we propose a simple yet effective framework that includes the hardness-weighted contrastive learning and the cross-device negative sample gathering strategy to enhance the model's learning of negative pairs with varying difficulty levels. This framework significantly improves the model’s capacity to distinguish between positive and negative pairs. Experimental results and in-depth analyses validate the effectiveness of the proposed framework. In the future, we plan to collect and construct a universal multimodal embedding benchmark for video-text retrieval, aiming to investigate the more universal multimodal embedding models. We will open-source all models and code, hoping to inspire further research in this field.
\section*{Limitations}
LLaVE is trained only on embedding datasets that contain arbitrary combinations of text and image modalities. Although it can generalize to embedding tasks that include the video modality in a zero-shot manner, there is still significant room for improvement. Constructing a multimodal embedding benchmark that incorporates the video modality will be a crucial direction for training more generalizable embedding models.

\section*{Acknowledgments}
The project was supported by 
National Key R\&D Program of China (No. 2022ZD0160501), 
Natural Science Foundation of Fujian Province of China (No. 2024J011001),
and
the Public Technology Service Platform Project of Xiamen (No.3502Z20231043).
We also thank the reviewers for their insightful comments.

\bibliography{acl_latex}

\newpage
\appendix
\section{Appendix}
\label{sec:appendix}

\subsection{Training Details}
\label{sec:training_details}
\begin{table}[t]
\centering
\resizebox{0.48\textwidth}{!}{
\begin{tabular}{l|ccc}
\toprule
Hyperparameter          & LLaVE-0.5B                          & LLaVE-2B                            & LLaVE-7B                           \\ \midrule
\#Data               & 662K                                & 662K                                & 662K                               \\
Batch size              & 256                                 & 256                                 & 256                                \\
lr                      & 1e-5                                & 1e-5                                & 5e-6                               \\
lr schedule             & \multicolumn{3}{c}{cosine decay}                                                                               \\
lr warmup ratio         & \multicolumn{3}{c}{0.03}                                                                                       \\
Weight decay            & \multicolumn{3}{c}{0}                                                                                          \\
Epoch                   & \multicolumn{3}{c}{1}                                                                                          \\
Optimizer               & \multicolumn{3}{c}{AdamW}                                                                                      \\
DeepSpeed stage         & \multicolumn{3}{c}{3}                                                                                          \\
Precision                      & Bf16 and TF32                                & Bf16 and TF32                                & Bf16                               \\
GPU                     & \multicolumn{1}{l}{8 $\times$ A100} & \multicolumn{1}{l}{8 $\times$ A100} & \multicolumn{1}{l}{16 $\times$ 910B} \\
Training cost (\#Hours) & 12                                  & 17                                  & 33                                 \\ \bottomrule
\end{tabular}
}
\caption{Training details of LLaVE.}
\label{tab:training_details}
\end{table}

We present the training details of LLaVE in Table \ref{tab:training_details}. To save memory and accelerate training, we adopt Gradient Checkpointing and Flash Attention \cite{DBLP:conf/nips/DaoFERR22} techniques. Due to resource constraints, LLaVE-7B is trained on 910B GPUs, and the training time will significantly decrease when conducted on A100 GPUs. In addition, using more GPUs to adopt a larger batch size or higher resolution will improve the model's performance.

\subsection{The Impact of Different Batch Sizes}
\begin{figure}[t]
    \centering
    \includegraphics[width=0.95\columnwidth]{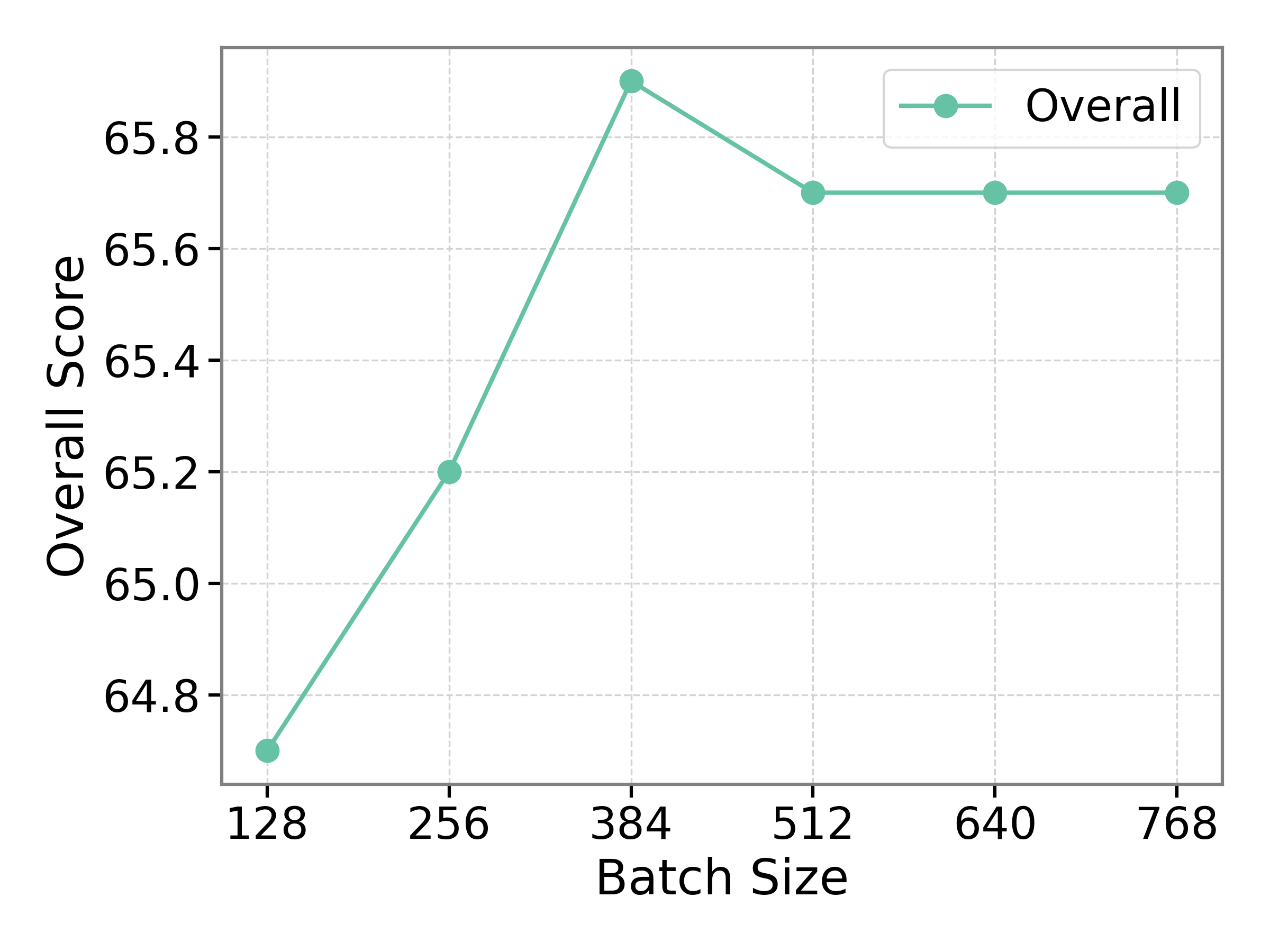}
    \vspace{-0.3cm}
    \caption{The impact of batch size on overall model performance when training for one epoch.}
    \label{figure:batch_size}
\end{figure} 
As shown in Figure \ref{figure:batch_size}, we also explore the impact of different batch sizes on the overall performance of the model when training for one epoch. It can be observed that as the batch size increases, the overall performance of the model improves. However, since the total number of trained epochs does not increase, larger batch sizes do not achieve further performance improvements. This is similar to the trend observed by \citeauthor{SigLIP} (\citeyear{SigLIP}), as large batch sizes need a sufficiently long schedule to ramp up. Besides, we find that the similarity distribution does not differ significantly across different batch sizes, which means that the similarity distribution remains relatively stable using different batch sizes.

\begin{figure}[t]
    \centering
    \includegraphics[width=0.88\columnwidth]{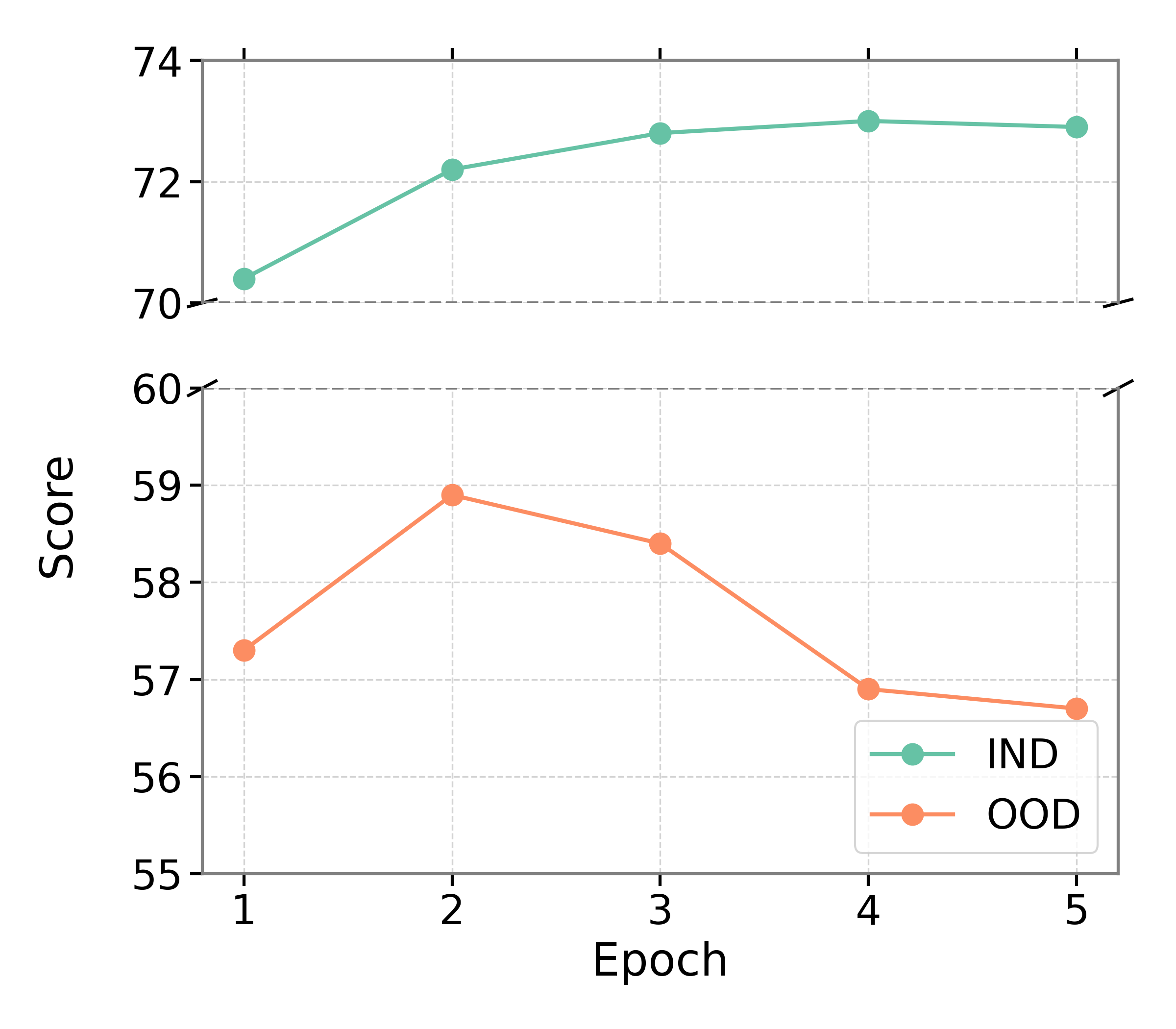}
    \vspace{-0.3cm}
    \caption{Impact of training epochs on IND and OOD performance.}
    \label{figure:epoch}
\end{figure} 

\subsection{The Impact of Training Epochs}
To further investigate the impact of training duration, we fine-tune LLaVE-2B for up to five epochs. The results in Figure \ref{figure:epoch} show that increasing the number of epochs initially enhances in-domain performance and overall accuracy. However, after three epochs, overfitting appears, which degrades out-of-domain performance and limits further overall improvements. For the sake of training efficiency and generalization, we therefore train for one epoch throughout this paper.

\begin{table*}[t]
\centering
\setlength\tabcolsep{8pt}%调列距
\renewcommand\arraystretch{1.2}
\small
\fontsize{8pt}{8pt}\selectfont
% \resizebox{\textwidth}{!}{
\begin{tabular}{lccccccc}
\toprule
\multicolumn{1}{c}{\multirow{2}{*}{\textbf{Model}}} & \multicolumn{4}{c}{\textbf{Per Meta-Task Score}}               & \multicolumn{3}{c}{\textbf{Average Score}}    \\ \cline{2-8} 
\multicolumn{1}{c}{}                                & Classification & VQA           & Retrieval     & Grounding     & IND           & OOD           & Overall       \\ \midrule
\# Datasets                                         & 10             & 10            & 12            & 4             & 20            & 16            & 36            \\ \midrule
VLM2Vec (Qwen2-VL-2B)                                & 59.0           & 49.4          & 65.4          & 73.4          & 66.0          & 52.6          & 60.1          \\
VLM2Vec (Qwen2-VL-7B)                                & 62.6           & 57.8          & \underline{69.9}          & \underline{81.7}          & \underline{72.2}          & \underline{57.8}          & \underline{65.8}          \\
\midrule
LLaVE-2B (Qwen2-VL-2B)                                           & \underline{64.3}           & \underline{58.5}          & 66.5          & 76.9          & 71.1          & 57.1          & 64.8          \\
LLaVE-7B (Qwen2-VL-7B)                                           & \textbf{65.4}  & \textbf{65.2} & \textbf{70.6} & \textbf{84.8} & \textbf{74.8} & \textbf{62.3} & \textbf{69.2} \\

\bottomrule
\end{tabular}
% }
\vspace{-0.2cm}
\caption{Comparison of LLaVE and VLM2Vec based on different LMMs.}
\label{tab:diff_lmms}
% \vskip -0.15in
\end{table*}

\subsection{Hyperparameter Analysis}
\label{sec:hyper_analysis}
\begin{figure}[t]
    \centering
    \includegraphics[width=0.85\columnwidth]{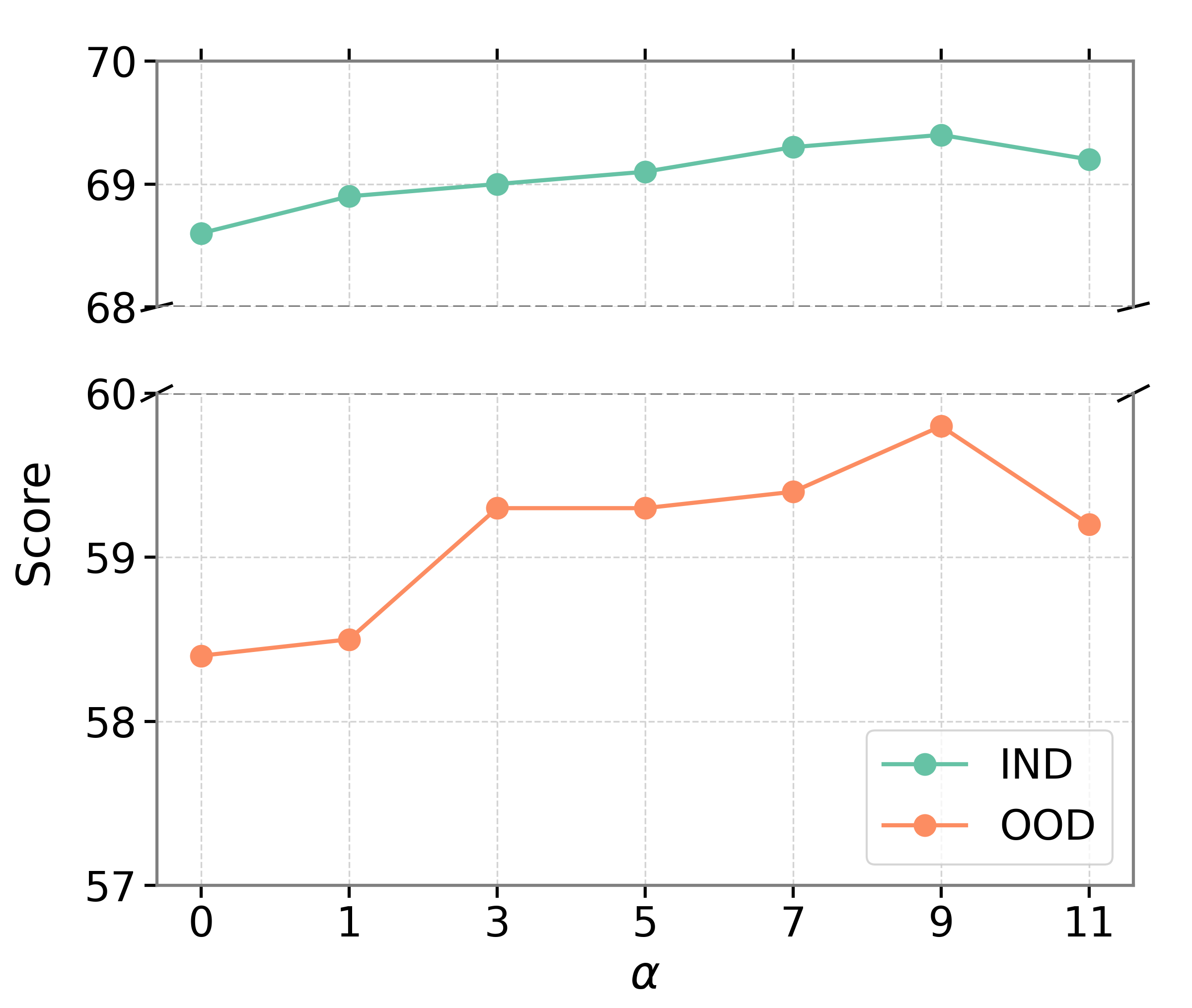}
    \vspace{-0.3cm}
    \caption{Influence of the $\alpha$ on model performance, measured on IND and OOD datasets, respectively.}
    \label{figure:alpha}
\end{figure} 

We experimentally explore the influence of the hyperparameter $\alpha$ on model performance. Figure \ref{figure:alpha} shows that increasing $\alpha$ positively influences both IND and OOD performance, stopping at a certain point. This indicates that appropriately weighting hard negative samples helps the model learn effectively. Moreover, it can be observed that the model's performance is robust to the $\alpha$ parameter, consistently outperforming the results obtained without hardness-weighted contrastive learning (i.e., when $\alpha$=0).

\subsection{Generalization of LLaVE Across LMMs}
To further validate the generality of our proposed methods, we extend LLaVE to other widely used LMMs and evaluated it on MMEB. Specifically, we apply our methods to Qwen2-VL models of different scales (2B and 7B) and employ the same training settings as VLM2Vec, including identical batch sizes and training steps.
To further validate the generalizability of our proposed methods, we extend LLaVE to other widely used LMMs and evaluate its performance on MMEB. Specifically, we apply our methods to Qwen2-VL models of varying scales (2B and 7B) and employ the same training settings as VLM2Vec. As shown in Table 5, when using the same LMM as the backbone, LLaVE still significantly outperforms VLM2Vec. In particular, LLaVE-2B achieves an improvement of 4.7 over VLM2Vec-2B, and LLaVE-7B achieves an improvement of 3.4 over VLM2Vec-7B. These results fully demonstrate the strong generalization capability of our method.

\subsection{Detailed results on MMEB}
\label{sec:detailed_result}
We present the detailed results of each model on various datasets of MMEB in Table \ref{tab:appendix_result}.

\begin{table*}[]
\centering
\resizebox{1.0\textwidth}{!}{
\begin{tabular}{lcccccccccc}
\toprule
  & \textbf{CLIP} & \textbf{OpenCLIP} & \textbf{SigLIP} & \textbf{BLIP2} & \textbf{MagicLens} & \textbf{E5-V} & \textbf{UniIR} & \textbf{VLM2Vec} & \textbf{MMRet} & \textbf{LLaVE}\\
\midrule
\rowcolor{orange!15} \textbf{Classification (10 tasks)} & & & & & & & & & &\\
ImageNet-1K          & 55.8 & 63.5 & 45.4 & 10.3 & 48.0 & 9.6 & 58.3 & 67.8 & 58.8 & 77.1\\
N24News              & 34.7 & 38.6 & 13.9 & 36.0 & 33.7 & 23.4 & 42.5 & 76.3 & 71.3 & 82.1\\
HatefulMemes         & 51.1 & 51.7 & 47.2 & 49.6 & 49.0 & 49.7 & 56.4 & 65.8 & 53.7 & 74.3\\
VOC2007              & 50.7 & 52.4 & 64.3 & 52.1 & 51.6 & 49.9 & 66.2 & 88.9 & 85.0 & 90.3\\
SUN397               & 43.4 & 68.8 & 39.6 & 34.5 & 57.0 & 33.1 & 63.2 & 74.4 & 70.0 & 79.1\\
\rowcolor{yellow!5} Place365  & 28.5 & 37.8 & 20.0 & 21.5 & 31.5 & 8.6 & 36.5 & 43.0 & 43.0 & 45.1\\
\rowcolor{yellow!10} ImageNet-A & 25.5 & 14.2 & 42.6 & 3.2  & 8.0  & 2.0 & 9.8 & 51.4 & 36.1 & 51.6\\
\rowcolor{yellow!10} ImageNet-R & 75.6 & 83.0 & 75.0 & 39.7 & 70.9 & 30.8 & 66.2 & 86.3 & 71.6 & 90.9\\
\rowcolor{yellow!10} ObjectNet  & 43.4 & 51.4 & 40.3 & 20.6 & 31.6 & 7.5 & 32.2& 59.5 & 55.8 & 46.2 \\
\rowcolor{yellow!10} Country-211 & 19.2 & 16.8 & 14.2 & 2.5  & 6.2  & 3.1 & 11.3 & 21.4 & 14.7 & 20.1 \\
\textit{All Classification} & 42.8 & 47.8 & 40.3 & 27.0 & 38.8 & 21.8 & 44.3 & 63.5 & 56.0 & 65.7\\
\midrule

\rowcolor{blue!15} \textbf{VQA (10 tasks)} & & & & & & & & & &\\
OK-VQA               & 7.5  & 11.5 & 2.4  & 8.7  & 12.7 & 8.9 & 25.4 & 67.3 & 73.3 & 71.1\\
A-OKVQA              & 3.8  & 3.3  & 1.5  & 3.2  & 2.9  & 5.9 & 8.8 & 63.6 & 56.7 & 70.8\\
DocVQA               & 4.0  & 5.3  & 4.2  & 2.6  & 3.0  & 1.7 & 6.2 & 86.6 & 78.5 & 90.3\\
InfographicsVQA      & 4.6  & 4.6  & 2.7  & 2.0  & 5.9  & 2.3 & 4.6 & 51.9 & 39.3 & 53.5\\
ChartQA              & 1.4  & 1.5  & 3.0  & 0.5  & 0.9  & 2.4 & 1.6 & 54.9 & 41.7 & 62.2\\
Visual7W             & 4.0  & 2.6  & 1.2  & 1.3  & 2.5  & 5.8 & 14.5 & 48.7 & 49.5 & 55.8\\
\rowcolor{yellow!10} ScienceQA  & 9.4  & 10.2 & 7.9  & 6.8  & 5.2  & 3.6 & 12.8 & 46.6 & 45.2 & 54.4\\
\rowcolor{yellow!10} VizWiz    & 8.2  & 6.6  & 2.3  & 4.0  & 1.7  & 2.6  & 24.3 & 48.3 & 51.7 & 48.5\\
\rowcolor{yellow!10} GQA        & 41.3 & 52.5 & 57.5 & 9.7  & 43.5 & 7.8 & 48.8 & 66.8 & 59.0 & 68.4\\
\rowcolor{yellow!10} TextVQA    & 7.0  & 10.9 & 1.0  & 3.3  & 4.6  & 8.2 & 15.1 & 76.0 & 79.0 & 79.4\\
\textit{All VQA}      & 9.1  & 10.9 & 8.4  & 4.2  & 8.3  & 4.9 & 16.2 & 61.1 & 57.4 & 65.4\\
\midrule

\rowcolor{green!15} \textbf{Retrieval (12 tasks)} & & & & & & & & & &\\
VisDial              & 30.7 & 25.4 & 21.5 & 18.0 & 24.8 & 9.2 & 42.2 & 73.8 & 83.0 & 83.0\\
CIRR                 & 12.6 & 15.4 & 15.1 & 9.8  & 39.1 & 6.1 & 51.3 & 50.3 & 61.4 & 54.5\\
VisualNews\_t2i      & 78.9 & 74.0 & 51.0 & 48.1 & 50.7 & 13.5 & 74.3 & 69.7 & 74.2 & 76.6\\
VisualNews\_i2t      & 79.6 & 78.0 & 52.4 & 13.5 & 21.1 & 8.1 & 76.8 & 72.3 & 78.1 & 81.2\\
MSCOCO\_t2i          & 59.5 & 63.6 & 58.3 & 53.7 & 54.1 & 20.7 & 68.5 & 74.5 & 78.6 & 78.9\\
MSCOCO\_i2t          & 57.7 & 62.1 & 55.0 & 20.3 & 40.0 & 14.0 & 72.1 & 71.7 & 72.4 & 74.7\\
NIGHTS               & 60.4 & 66.1 & 62.9 & 56.5 & 58.1  & 4.2 & 66.2 & 66.5 & 68.3 & 67.0\\
WebQA                & 67.5 & 62.1 & 58.1 & 55.4 & 43.0 & 17.7 & 89.6 & 87.5 & 90.2 & 90.4\\
\rowcolor{yellow!10} FashionIQ  & 11.4 & 13.8 & 20.1 & 9.3  & 11.2 & 2.8 & 40.2 & 20.6 & 54.9 & 23.3\\
\rowcolor{yellow!10} Wiki-SS-NQ & 55.0 & 44.6 & 55.1 & 28.7 & 18.7 & 8.6 & 12.2 & 53.7 & 24.9 & 63.9\\
\rowcolor{yellow!10} OVEN       & 41.1 & 45.0 & 56.0 & 39.5 & 1.6  & 5.9 & 69.4 & 67.0 & 87.5 & 68.0\\
\rowcolor{yellow!10} EDIS       & 81.0 & 77.5 & 23.6 & 54.4 & 62.6 & 26.8 & 79.2 & 66.6 & 65.6 & 89.1\\
\textit{All Retrieval} & 53.0 & 52.3 & 31.6 & 33.9 & 35.4 & 11.5 & 61.8 & 64.5 & 69.9 & 70.9\\
\midrule

\rowcolor{purple!15} \textbf{Visual Grounding (4 tasks)} & & & & & & & & & &\\
MSCOCO         & 33.8 & 34.5 & 46.4 & 28.9 & 22.1 & 10.8 & 46.6 & 80.6 & 76.8 & 87.0\\
\rowcolor{yellow!10} RefCOCO  & 56.9 & 54.2 & 70.8 & 47.4 & 22.8 & 11.9 & 67.8 & 92.3 & 89.8 & 95.4\\
\rowcolor{yellow!10} RefCOCO-matching  & 61.3 & 68.3 & 50.8 & 59.5 & 35.6 & 38.9 & 62.9 & 85.3 & 90.6 & 92.8\\
\rowcolor{yellow!10} Visual7W-pointing & 55.1 & 56.3 & 70.1 & 52.0 & 23.4 & 14.3 & 71.3 & 91.0 & 77.0 & 92.5\\
\textit{All Visual Grounding} & 51.8 & 53.3 & 59.5 & 47.0 & 26.0 & 19.0 & 65.3 & 87.3 & 83.6 & 91.9\\
\midrule

\rowcolor{cyan!15} \textbf{Final Score (36 tasks)} & & & & &  & & & & &\\
All                  & 37.8 & 39.7 & 34.8 & 25.2  & 27.8 & 13.3 & 44.7 & 65.8 & 64.1 & 70.3\\
All IND       & 37.1 & 39.3 & 32.3 & 25.3  & 31.0 & 14.9 & 47.1 & 61.0 & 59.1 & 64.4\\
All OOD       & 38.7 & 40.2 & 38.0 & 25.1  & 23.7 &  11.5 & 41.7 & 69.7 & 68.0 & 75.0\\

\bottomrule
\end{tabular}
}
\caption{The detailed results of the baselines and LLaVE on MMEB. The out-of-distribution datasets are highlighted with a yellow background in the table. We only include the best version of each series of models in the table, such as LLaVE-7B and VLM2Vec (LLaVA-OV-7B).}
\label{tab:appendix_result}
\end{table*}

\end{document}